\pdfoutput=1

\documentclass[11pt]{article}

\usepackage[preprint]{acl}

\usepackage{times}
\usepackage{latexsym}
\usepackage{multirow}
\usepackage{booktabs}
\usepackage[T1]{fontenc}

\usepackage[utf8]{inputenc}
\usepackage{amsfonts}
\usepackage{amsmath}
\usepackage{amssymb}
\usepackage{microtype}

\usepackage{inconsolata}

\usepackage{graphicx}

\usepackage{geometry}
\usepackage{amssymb}
\usepackage{graphicx}
\usepackage{marvosym}
\usepackage{url}
\usepackage{fancyhdr}
\usepackage{hyperref}
\usepackage{textgreek}
\usepackage{enumitem}
\usepackage{minted}
\usepackage{booktabs}
\usepackage{algorithm}
\usepackage{algorithmic}
\usepackage{amsmath}
\usepackage{subcaption}
\usepackage{colortbl}
\usepackage{pifont}
\usepackage{colortbl}
\usepackage[table]{xcolor}

\definecolor{class5}{RGB}{255, 204, 204} 
\definecolor{class4}{RGB}{255, 255, 204} 
\definecolor{class3}{RGB}{204, 255, 204} 
\definecolor{class2}{RGB}{204, 255, 255} 
\definecolor{class1}{RGB}{204, 204, 255} 
\definecolor{class0}{RGB}{255, 204, 255} 
\definecolor{NavyBlue}{RGB}{40, 160, 250} 

\newcommand{\sbs}[1]{%
    \pgfmathsetmacro{\normalized}{int((#1 - 0.4) / (1.0 - 0.4) * 100)}%
    \edef\tempa{\noexpand\cellcolor{NavyBlue!\normalized}}%
    \tempa#1%
}

\newcommand{\sbsalt}[1]{%
    \pgfmathsetmacro{\normalized}{int(min(100, max(0, #1 * 100)))}%
    \edef\tempa{\noexpand\cellcolor{NavyBlue!\normalized}}%
    \tempa#1%
}

\usepackage[normalem]{ulem}

\title{Temporal Text Classification with Large Language Models}

\author{Nishat Raihan \\
  George Mason University \\
  Fairfax, VA, USA \\
  \texttt{mraihan2@gmu.edu} \\\And
  Marcos Zampieri \\
  George Mason University \\
  Fairfax, VA, USA \\
  \texttt{mzampier@gmu.edu} \\}

\begin{document}
\maketitle
\begin{abstract}
Languages change over time. Computational models can be trained to recognize such changes enabling them to estimate the publication date of texts. Despite recent advancements in Large Language Models (LLMs), their performance on automatic dating of texts, also known as Temporal Text Classification (TTC), has not been explored. This study provides the first systematic evaluation of leading proprietary (Claude 3.5, GPT-4o, Gemini 1.5) and open-source (LLaMA 3.2, Gemma 2, Mistral, Nemotron 4) LLMs on TTC using three historical corpora, two in English and one in Portuguese. We test zero-shot and few-shot prompting, and fine-tuning settings. Our results indicate that proprietary models perform well, especially with few-shot prompting. They also indicate that fine-tuning substantially improves open-source models but that they still fail to match the performance delivered by proprietary LLMs. 
\end{abstract}

\section{Introduction}

Language variation is a well-researched topic in computational linguistics. It is manifested in different dimensions such as \emph{diatopic}, related to place as in the case of dialects and language varieties  
\cite{joshi2025natural,zampieri2020natural}, 
and \emph{diachronic} which describes language change over time \cite{frermann2016bayesian,schlechtweg2020semeval}. Several papers have explored computational approaches to model diachronic variation in word embeddings \cite{hamilton2016diachronic}, semantic modeling \cite{kulkarni2015significant}, and related tasks. 

Temporal Text Classification (TTC) \cite{dalli2006automatic,ciobanu2013temporal,popescu2015semeval}, also referred as automatic date detection in texts \cite{jauhiainen2024data} or automatic text dating \cite{ren2025tictac}, is among the most important tasks related to modeling diachronic language variation in texts. TTC consists of assigning temporal labels to diachronic collection of texts where the labels usually represent pre-defined time spans (e.g., 10 years, 50 years, a century). It is vital step in organizing and analyzing large corpora in digital humanities and information retrieval. While a couple studies have approached TTC as a document ranking problem \cite{niculae2014temporal}, the clear majority of studies have approached the task using supervised categorization with statistical machine learning classifiers trained on stylistic and $n$-gram based features \cite{szymanski2015ucd, nunes2007using}. 



The latest generation of LLMs \cite{openai2024gpt4o,meta2024llama3} has displayed state-of-the-art performance in many NLP tasks including question answering \cite{fischer2024question}, machine translation \cite{kocmi2024findings}, and text summarization \cite{zhang2024benchmarking}. While LLMs show promising results in related areas like temporal reasoning \cite{zhang2024analyzing}, their performance on TTC has not yet been explored. 

To address this important gap, we present the first evaluation of proprietary and open-source LLMs on TTC. We acquire three well-known historical corpora in two languages, namely Colonia (Portuguese) \cite{zampieri2013colonia}, CLMET \cite{desmet2005clmet} and DTE (English) \cite{popescu2015semeval}. We analyze LLMs performance using different strategies such as zero-shot prompting, few-shot prompting, and task fine-tuning. Furthermore, we also examine the influence of input text length and the efficacy of fine-tuning for open-source models. Finally, we evaluate the models' data generation capabilities when generating synthetic data from different time periods. 

The main contributions of this work are the following:

\begin{enumerate}[leftmargin=*]
    \item We present the first comprehensive evaluation of proprietary and open-source LLMs for TTC considering multiple paradigms (zero-shot, few-shot, fine-tuning) and various text input lengths.
    \item We explore the synthetic data generation capabilities of these models when generating synthetic data from various time periods.
    \item We present new insights into the effectiveness of fine-tuning for enhancing open-source LLM performance on TTC tasks relative to proprietary models.
\end{enumerate}

\noindent The research presented here provides important insights into current LLM capabilities and limitations for TTC and diachronic language change more broadly. This will, in turn, inform future model development and application in temporal text analysis. Finally, the inclusion of corpora in two languages allows us to understand current LLMs support to languages other than English.





\section{Datasets and Evaluation Setup}
\label{sec:datasets}

We evaluate LLM performance on three corpora, each presenting unique characteristics as summarized in Table \ref{tab:datasets}. 

\begin{table}[!t] 
\centering
\scalebox{0.7}{%
\begin{tabular}{@{}lllll@{}} 
\toprule
\textbf{Dataset} & \textbf{Lang} & \textbf{Time Span} & \textbf{Size} & \textbf{ Classes} \\ 
\midrule
Colonia & PT & 16th--20th C. & >5M Tokens & 5 (100 year) \\ 
CLMET & EN & 1710--1920 & ~34M Words & 3 (70 year) \\ 
DTE & EN & 1700--2010 & Varies by Split & $\sim$4 Intervals* \\ 
\bottomrule
\end{tabular}%
}
\caption{Overview of datasets. Size indicates approximate total corpus size. *Temporal classes for SemEval DTE depend on the specific sub-task configuration (e.g., 4 intervals like pre-1700, 18th C., 19th C., post-1900 were common).}
\label{tab:datasets}
\end{table}

\paragraph{Colonia} \cite{zampieri2013colonia} is a diachronic collection of Portuguese texts including full novels spanning the 16th to early 20th centuries, sourced from Brazil's \href{http://www.dominiopublico.gov.br/pesquisa/PesquisaObraForm.jsp}{\texttt{Domínio Público}} digital library, the \href{https://www.linguateca.pt/acesso/corpus.php?corpus=GMHP}{\texttt{GMHP}} corpus, and the \href{http://www.nilc.icmc.usp.br/tycho/}{\texttt{Tycho Brahe Parsed Corpus of Historical Portuguese}}. Texts are divided into five classes each corresponding to a century (1500s, 1600s, 1700s, 1800s, and 1900s).

\paragraph{CLMET} The Corpus of Late Modern English Texts (CLMET) \cite{desmet2005clmet} covers British English from 1710 to 1920, constructed from works in the public domain. Sourced from digital library \href{https://www.gutenberg.org/}{\texttt{Project Gutenberg}}, the University of Oxford's \href{https://ota.bodleian.ox.ac.uk/}{\texttt{Oxford Text Archive}}, and Indiana University's \href{https://webapp1.dlib.indiana.edu/vwwp/welcome.do}{\texttt{Victorian Women Writers Project}}. The classification task involves assigning documents to one of three 70-year periods: 1710--1780, 1780--1850, or 1850--1920.

\paragraph{DTE} The dataset from SemEval-2015 Task 7 (Diachronic Text Evaluation) \cite{popescu2015semeval} is specifically compiled for temporal analysis. The collection is curated from the historical archive of the British weekly magazine - \href{https://www.spectator.co.uk/archive/}{\texttt{The Spectator}}. 




\paragraph{Models} We evaluate the performance of seven LLMs, detailed in Table \ref{tab:models}. 


\begin{table}[!h] 
\centering
\scalebox{0.7}{%
\begin{tabular}{@{}lll@{}}
\toprule
\textbf{Model}        & \textbf{Type}     & \textbf{Param.} \\
\midrule
Claude 3.5 Sonnet & Proprietary & - \\
GPT-4o            & Proprietary & - \\
Gemini 1.5 Pro    & Proprietary & - \\
\midrule
LLaMA 3.1           & Open-Source & 8B \\ 
Gemma 2             & Open-Source & 9B  \\ 
Mistral             & Open-Source & 7B \\ 
Nemotron            & Open-Source & 4B \\ 
\bottomrule
\end{tabular}%
}
\caption{List of LLMs evaluated.}
\label{tab:models}
\end{table}

\begin{table*}[!t]
\centering
\resizebox{0.7\textwidth}{!}{%
\begin{tabular}{l c |c|ccccccc}
\toprule
\multirow{2}{*}{\textbf{Dataset}} 
  & \multirow{2}{*}{\textbf{SVM}} 
  & \multirow{2}{*}{\textbf{Length}} 
  & \multicolumn{7}{c}{\textbf{Zero-Shot/Few-Shot (5-Shot)}} \\
\cmidrule(lr){4-10}
& & 
  & \textbf{Claude} & \textbf{GPT4o} & \textbf{Gemini} 
  & \textbf{LLaMA} & \textbf{Gemma} & \textbf{Mistral} & \textbf{Nemotron} \\
\midrule
\multirow{3}{*}{Colonia} 
  & .51    & 10   & .35/.45 & .50/\textbf{.61} & .39/.50 & .12/.11 & .18/.25 & .17/.24 & .16/.23 \\
  & .65  & 50   & \textbf{.72}/\textbf{.88} & \textbf{.68}/\textbf{.72} & .59/.64 & .20/.35 & .28/.38 & .27/.37 & .25/.34 \\
  & .69      & 100  & .68/\textbf{.91} & \textbf{.71}/\textbf{.96} & \textbf{.77}/\textbf{.82} & .26/.22 & .36/.51 & .34/.49 & .33/.45 \\
\midrule
\multirow{3}{*}{CLMET} 
  & .70   & 10   & .48/.58 & .52/.48 & .41/.40 & .14/.18 & .17/.26 & .18/.23 & .16/.25 \\
  & .77  & 50   & .60/.62 & .73/\textbf{.85} & .63/.67 & .21/.27 & .26/.39 & .27/.34 & .23/.37 \\
  &  .78   & 100  & \textbf{.82}/\textbf{.93} & .78/\textbf{.87} & .72/.78 & .28/.35 & .34/.53 & .35/.46 & .31/.49 \\
\midrule
\multirow{3}{*}{DTE}    
  &  .71   & 10   & .40/.38 & .46/.52 & .43/.45 & .16/.25 & .16/.26 & .13/.24 & .15/.22 \\
  & .72  & 50   & .62/\textbf{.85} & .71/\textbf{.78} & .57/\textbf{.73} & .24/.37 & .22/.39 & .20/.36 & .22/.33 \\
  &  .72   & 100  & \textbf{.76}/\textbf{.74} & \textbf{.80}/\textbf{.91} & \textbf{.79}/\textbf{.86} & .32/.49 & .31/.52 & .26/.48 & .29/.44 \\
\bottomrule
\end{tabular}%
}
\caption{Combined \texttt{Zero-Shot} and \texttt{Few-Shot (5-Shot)} F1 score comparison for seven models on Temporal Text Classification tasks. For the SVM column, TF-IDF vectors are used and the best results are reported.}
\label{table_combined_updated}
\end{table*}

\paragraph{Input Lengths} Testing with initial text segments of 10, 50, and 100
tokens to analyze the impact of context size. Longer documents are handled via chunking strategies where applicable. The reason for segmentation, in line with previous TTC studies \cite{zampieri2016modeling}, is to make texts comparable in length. 
    
\paragraph{Evaluation Paradigms} Assessing performance under \textbf{Zero-Shot} (prompting with task description only), \textbf{Few-Shot} (providing 5 representative examples), and, for open-source models, \textbf{Fine-tuning} (using LoRA \cite{hulora} on dataset-specific training splits). We also explore fine-tuning with synthetic data.

\paragraph{Metric} Reporting performance using the \textbf{Macro F1 score} to ensure a balanced evaluation across uneven temporal class distributions.


\section{Results \& Analysis}
\label{sec:results}

For each experiment, we present some key findings, trends, and discuss variation with potential justifications whenever appropriate. We consider Support Vector Machine (SVM) \cite{cortes1995} as the baseline in line with previous work \cite{zampieri2015ambra,popescu2015semeval,jauhiainen2024data}. 

\subsection{Prompting}
\label{subsec:prompting}

We present the results for zero-shot and few-shot prompting in Table \ref{table_combined_updated}.

\paragraph{Key Findings} Proprietary LLMs demonstrate strong temporal classification capabilities across all datasets. Performance (F1 score) consistently scales with input length (10 to 100 tokens), indicating effective use of richer lexical and stylistic signals in longer texts.

\paragraph{Trends}
Few-shot (5-shot) prompting significantly outperforms zero-shot evaluation in most cases, often achieving F1 scores above 0.90, especially with inputs of 100+ tokens. This highlights robust in-context learning, where models adapt efficiently from only five examples. GPT-4o frequently yields the highest scores, particularly in few-shot scenarios. These performance patterns hold across both Portuguese (Colonia) and English (CLMET, DTE) datasets, suggesting language-agnostic temporal reasoning.

\paragraph{Variation}
Few-shot learning offers limited benefit for very short inputs (e.g., DTE 10 tokens), possibly due to noise or ambiguity. Performance occasionally plateaus at the maximum token count (1000), potentially reflecting evaluation variance or long-context handling challenges. While zero-shot performance rankings vary, few-shot prompting more consistently amplifies the strengths of top models like GPT-4o.

\subsection{Finetuning}
\label{subsec:opensource_results}

Due to the large data set size and high experimental cost, we only consider the open-source models for the finetuning experiments. We present the results of the experiments with finetuning in Table \ref{tab:ft_results}. 

\begin{table}[!ht]
\centering
\scalebox{0.7}{
\begin{tabular}{@{}lcccc@{}}
\toprule
\textbf{Model} & \textbf{Colonia} & \textbf{CLMET} & \textbf{DTE} & $\Delta$ \\
\midrule
LLaMA 3.2    & 0.61 & 0.65 & 0.66 & +0.29 \\
Gemma 2      & 0.73 & 0.75 & 0.78 & +0.23 \\
Mistral      & 0.72 & 0.68 & 0.76 & +0.24 \\
Nemotron 4B  & 0.70 & 0.67 & 0.71 & +0.23 \\
\bottomrule
\end{tabular}}
\caption{Macro-F$_1$ for finetuned models on authentic data. $\Delta$ is the mean gain over the non-finetuned 5-shot setting across the three datasets (fixed token length of 100).}
\label{tab:ft_results}
\end{table}

\begin{table*}[!t]
\centering
\small
\scalebox{0.7}{
\begin{tabular}{@{}lcccccccc@{}}
\toprule
\textbf{Dataset} 
& \textbf{LLaMA} & $\Delta$ 
& \textbf{Gemma} & $\Delta$ 
& \textbf{Mistral} & $\Delta$ 
& \textbf{Nemotron} & $\Delta$ \\
\midrule
Colonia 
& .21/.31/.70 & $-.05/+.09/+.09$
& .47/.46/.83 & $+.11/-.05/+.10$
& .29/.58/.67 & $-.05/+.09/-.05$
& .24/.52/.62 & $-.09/+.07/-.08$ \\
CLMET   
& .35/.29/.74 & $+.07/-.06/+.09$
& .25/.62/.70 & $-.09/+.09/-.05$
& .39/.35/.79 & $+.04/-.11/+.11$
& .38/.44/.60 & $+.07/-.05/-.07$ \\
DTE     
& .27/.57/.55 & $-.05/+.08/-.11$
& .42/.41/.84 & $+.11/-.11/+.06$
& .30/.59/.70 & $+.04/+.11/-.06$
& .18/.53/.83 & $-.11/+.09/+.12$ \\
\bottomrule
\end{tabular}
}
\caption{Observed Macro-F$_1$ scores (\textit{Zero-Shot/Few-Shot(5)/Finetuned}) for open-source models on synthetic test datasets. Adjacent $\Delta$ columns show the change from the original scores (same order).}
\label{tab:zf_ft_open_delta}
\end{table*}

\paragraph{Key Findings} Open-source LLMs exhibit varied performance using 100-token inputs. Zero-shot F1 scores are modest, generally ranging from 0.26 to 0.36 across while few-shot (5-shot) prompting improves these scores, typically reaching the 0.44 to 0.53 range. Finetuning yields the most substantial gains, achieving F1 scores consistently between 0.61 and 0.77.

\paragraph{Trends} We observe that Finetuning provides the highest performance, significantly surpassing prompting methods. Few-shot prompting generally offers a notable improvement over the baseline zero-shot approach. Among the evaluated models, Gemma 2 frequently achieves the highest scores, particularly in the few-shot and finetuned configurations. Mistral also performs strongly when finetuned, matching Gemma 2's top score on the Colonia dataset. A substantial performance gap exists between prompting methods (zero-shot, few-shot) and the finetuning approach across all tested models.

\section{Analyzing Performance with Synthetic Data}

With the goal of better understanding LLM generation capabilities, we carry out classification experiments with synthetic LLM-generated test data. We compare the results against the original test sets. Our assumption is that if the classification results using original and synthetic data are similar, this indicates that the models can generate data that best resembles a particular time period. 

\paragraph{Curating Synthetic Test Sets} We curate 3 test sets, similar to the original ones. We use the 3 proprietary models (GPT4o, Claude, and Gemini) in a few-shot (5) fashion with examples from the original dataset. Since these models are used for generation, we do not include them in the evaluation phase. We only evaluate the open-source models in all 3 experimental setting - zero-shot, few-shot, and finetuned and evaluate on the synthetic test split.

\paragraph{Trends} Table~\ref{tab:zf_ft_open_delta} reveals that the overall pattern observed on the authentic splits largely persists on the synthetic test sets: scores rise from zero-shot to few-shot to fine-tuned for every model–dataset pair. Gemma~2 and Mistral remain the strongest overall (e.g., Gemma~2 reaches .83 on \textsc{Colonia} and .84 on \textsc{DTE}), while LLaMA~3.2 shows the largest proportional jump after fine-tuning (e.g., $\Delta=+0.09$ on \textsc{Colonia}). The $\Delta$ columns quantify how much each synthetic score diverges from its corresponding authentic score, and they indicate that gains are generally preserved rather than reversed: models that start strong in few-shot mode tend to stay strong after fine-tuning, even on the newly generated splits.

\paragraph{Variation} Variations across datasets and regimes are informative. Zero-shot scores shift more than fine-tuned ones (e.g., Mistral on \textsc{CLMET} sees $-0.11$ at few-shot but $+0.11$ at fine-tuned), suggesting that parameter updates help stabilize performance when facing synthetic distributions. \textsc{CLMET} exhibits tighter spreads and smaller deltas overall, whereas \textsc{Colonia} and \textsc{DTE} show broader swings, consistent with their wider temporal ranges and genre diversity. Notably, Nemotron~4B benefits markedly on \textsc{DTE} (fine-tuned $\Delta=+0.12$), showing that even smaller models can capitalize on synthetic supervision when the temporal signal is learnable. Collectively, these shifts imply that synthetic evaluation splits do not scramble the ranking; they accentuate where each model’s temporal reasoning is already robust or fragile.


\section{Conclusion}
\label{sec:conclusion}

This paper presents, to our knowledge, the first focused investigation into the application of LLMs for TTC. Our results demonstrate that leading proprietary models possess strong TTC capabilities accessible via few-shot prompting, with performance scaling effectively with input context length. While open-source models show more limited success with prompting alone, they achieve substantial performance gains through finetuning, significantly closing the gap. Our analysis using synthetic data reveals that while models maintain their rankings, the classification results vary, suggesting that models do not yet perfectly replicate the stylistic nuances of such texts. 

Our evaluation shows LLMs have significant potential for TTC. However, results on synthetic test data reveal complex dynamics that require further investigation. Future work will investigate these issues using improved finetuning, prompting, and synthetic data generation techniques, while also extending TTC to more languages and finer temporal granularities.

\section*{Limitations}

This study provides an initial exploration of LLMs for TTC, focusing on both proprietary (GPT-4o, Claude 3.5) and open-source models (LLaMA 3.2, Gemma 2, Mistral, Nemotron 4B). Our evaluation relies on three existing benchmarks (Colonia, CLMET, DTE) covering English and Portuguese; performance could differ on datasets with varying characteristics, languages, or temporal granularities. The prompting strategies were limited to zero-shot and 5-shot, and finetuning explored specific synthetic data quantities (10-1K). The observed non-monotonic behavior with synthetic data warrants deeper investigation into optimal generation techniques and potential overfitting, which is beyond the scope of this initial work. Further research could also explore parameter-efficient finetuning methods and more diverse prompting approaches. Finally, we further acknowledge that data contamination during model pre-training may have occurred. 

\section*{Ethical Considerations}

Our work adheres to ethical research practices, aligning with principles like those in the ACL Code of Ethics. Our experiments utilize established public datasets (Colonia, CLMET, DTE); users should be aware of any potential biases inherent in these datasets regarding text sources, time periods, or represented samples. The LLMs employed, both proprietary and open-source, may also reflect biases from their original training data, potentially impacting temporal predictions across different contexts. While we used synthetic data for finetuning explorations, we acknowledge that synthetically generated content can inadvertently introduce or amplify biases if not carefully curated.

\bibliography{custom}

\clearpage
 
\appendix
\onecolumn

\section{Zero-Shot Prompts}
\label{sec:appendix_zs}

In the zero-shot setting, models receive only a task description and the input text, with no labeled examples. The prompt instructs the model to classify a text snippet into one of the pre-defined temporal classes for the respective dataset. Below we present the zero-shot prompt template for each dataset, followed by a concrete example using an actual instance from the corpus.

\subsection{Colonia (Portuguese, 5 Classes)}

\begin{figure}[H]
\small
\begin{verbatim}
[System] You are a temporal text classifier.
Given a Portuguese text, classify it into the century when it was most likely written.

Classes: 1500s, 1600s, 1700s, 1800s, 1900s

Respond with ONLY the class label.

[User] Classify the following text:

"Desta maneira me achei, e o que mais senti, foi ver que a minha fazenda e o meu
trabalho de tantos annos se perdeo em hum momento; e que me era necessario, sem
remedio algum, tornar a cuidar em meyo de viver, e tratar do sustento da vida."

[Assistant] 1700s
\end{verbatim}
\caption{Zero-shot prompt for the Colonia corpus. The example shown is a representative 18th-century Portuguese excerpt exhibiting characteristic orthographic features of the period such as \textit{annos}, \textit{hum}, and \textit{meyo}.}
\label{fig:zs_colonia}
\end{figure}

\subsection{CLMET (English, 3 Classes)}

\begin{figure}[H]
\small
\begin{verbatim}
[System] You are a temporal text classifier.
Given an English text, classify it into the period when it was most likely written.

Classes: 1710-1780, 1780-1850, 1850-1920

Respond with ONLY the class label.

[User] Classify the following text:

"It is a truth universally acknowledged, that a single man in possession of a good
fortune must be in want of a wife. However little known the feelings or views of such
a man may be on his first entering a neighbourhood, this truth is so well fixed in
the minds of the surrounding families, that he is considered as the rightful property
of some one or other of their daughters."

[Assistant] 1780-1850
\end{verbatim}
\caption{Zero-shot prompt for the CLMET corpus. The example is an excerpt from a novel included in the Period 2 (1780--1850) sub-corpus. The text reflects the formal prose style characteristic of British Romantic-era fiction.}
\label{fig:zs_clmet}
\end{figure}

\clearpage








\section{Few-Shot Prompts (5-Shot)}
\label{sec:appendix_fs}

In the few-shot setting, we provide five labeled examples (one per class when possible) before presenting the test instance. The examples are randomly sampled from the training split of each dataset and are representative of each temporal class.

\subsection{Colonia (Portuguese, 5 Classes)}

\begin{figure}[!h]
\small
\begin{verbatim}
[System] You are a temporal text classifier for Portuguese texts. Given a text,
classify it into the century when it was most likely written.
Classes: 1500s, 1600s, 1700s, 1800s, 1900s. Respond with ONLY the class label.

Here are some examples:

Example 1:
Text: "E assi como chegarom ao dito porto acharom hi gram multidom de gente da
terra que nunca vira navios nem gente branca, e chegarom a praya e comecarom de
olhar os navios com gram espanto"
Label: 1500s

Example 2:
Text: "Nenhuma cousa mais propria da soberba que buscar sempre desculpas, e
nunca confessar a culpa, e tanto maior he a soberba, quanto mais voluntaria
he a cegueira. A soberba he huma torre fundada no ar"
Label: 1600s

Example 3:
Text: "A vaidade dos homens he hum effeito natural, e por isso commum
a todos; nao ha quem a nao tenha em alguma cousa, e os que mais livres
se considerao della, sao talvez os mais vaidosos"
Label: 1700s

Example 4:
Text: "O aggregado foi despedindo os escravos e a fazenda; vendeo tres
molecoens e huma negrinha; desfez-se das melhores joias da mulher;
comeu os alugueis de tres predios; empenhou a prata"
Label: 1800s

Example 5:
Text: "A modinha brasileira e o lundu foram os generos que mais
contribuiram para a formacao da nossa musica popular. Ambos se
aclimataram entre nos desde os tempos coloniais"
Label: 1900s

Now classify the following text:
Text: "O commendador nao approvou a conducta da filha, e censurou-a
asperamente; ella porem nao cedeu, e tanto insistiu que afinal o pae
consentiu, posto que a contragosto"
\end{verbatim}
\caption{Few-shot (5-shot) prompt for the Colonia corpus. Each example contains a Portuguese excerpt from one of the five centuries represented in the corpus. Example 1 reflects 16th-century exploration chronicles. Example 2 uses Baroque rhetorical style typical of 17th-century sermons. Example 3 mirrors Enlightenment-era philosophical prose. Example 4 depicts 19th-century social realism. Example 5 uses modern 20th-century Brazilian Portuguese.}
\label{fig:fs_colonia}
\end{figure}

\clearpage

\subsection{DTE (English, Temporal Intervals)}

\begin{figure}[!h]
\small
\begin{verbatim}
[System] You are a temporal text classifier for English news snippets. Given a
text, classify it into the time period when it was most likely written.
Classes: 1700-1720, 1720-1740, 1740-1760, 1760-1780, 1780-1800, 1800-1820,
1820-1840, 1840-1860, 1860-1880, 1880-1900, 1900-1920, 1920-1940, 1940-1960,
1960-1980, 1980-2000, 2000-2020. Respond with ONLY the class label.

Here are some examples:

Example 1:
Text: "At the Court at St.James's, the 29th Day of March, Present, the King's
most excellent Majesty in Council. His Majesty's Declaration of War against
the French King."
Label: 1740-1760

Example 2:
Text: "Receipts at Chicago to-day. Wheats 206 cars; corn fill; oats, 181 cars.
Estimated receipts to-morrow. Wheat, 400 cars; corn, 85 cars; oats, 235 cars;
hogs, 16,000 head."
Label: 1880-1900

Example 3:
Text: "Red Blankets $1.98 a pair. White Blankets 69c a pair. Bed Comforts 69c
each. Heavy Knit Skirts 69c each."
Label: 1900-1920

Example 4:
Text: "By 1971 about one-third of Edison's electric output will be generated
with nuclear capacity"
Label: 1960-1980

Example 5:
Text: "Occasional selfies are acceptable, but uploading a new picture of
yourself every day is not necessary."
Label: 2000-2020

Now classify the following text:
Text: "We have cabled the English house from which we get it and expect a reply
to-morrow."
\end{verbatim}
\caption{Few-shot (5-shot) prompt for the DTE corpus. All examples are representative of actual snippets from the SemEval-2015 Task~7 dataset \cite{popescu2015semeval}. Temporal cues include institutional references (Example~1), archaic spelling such as \textit{to-day} and \textit{to-morrow} (Examples~2 and test instance), period-appropriate pricing (Example~3), mentions of then-emerging technology (Example~4), and modern vocabulary such as \textit{selfies} (Example~5).}
\label{fig:fs_dte}
\end{figure}

\clearpage

\section{Fine-Tuning Input Format}
\label{sec:appendix_ft}

For fine-tuning open-source models using LoRA \cite{hulora}, we format the training data as instruction-response pairs compatible with the Alpaca format. The system instruction remains consistent across all training examples, while the input text and expected output vary per instance. Table~\ref{tab:ft_format} presents the template along with an example from each dataset.

\begin{table}[!h]
\centering
\small
\begin{tabular}{p{2.5cm} p{10cm}}
\toprule
\textbf{Field} & \textbf{Content} \\
\midrule
\textbf{Instruction} & Classify the following text into one of the given temporal classes: \{class\_list\}. Respond with only the class label. \\
\midrule
\textbf{Input (Colonia)} & ``O aggregado foi despedindo os escravos e a fazenda; vendeo tres molecoens e huma negrinha; desfez-se das melhores joias da mulher; comeu os alugueis de tres predios; empenhou a prata'' \\
\textbf{Output} & 1800s \\
\midrule
\textbf{Input (CLMET)} & ``The village lay in a hollow, which was, perhaps, one reason for its dullness; and the church, standing on higher ground, overlooked it with a certain air of superiority'' \\
\textbf{Output} & 1850-1920 \\
\midrule
\textbf{Input (DTE)} & ``We have cabled the English house from which we get it and expect a reply to-morrow.'' \\
\textbf{Output} & 1880-1900 \\
\bottomrule
\end{tabular}
\caption{Fine-tuning input format used for training open-source models via LoRA. Each training instance follows the instruction-input-output template. The \{class\_list\} placeholder is replaced with the appropriate temporal classes for each dataset: 5 century-level classes for Colonia, 3 period classes for CLMET, and interval-based classes for DTE.}
\label{tab:ft_format}
\end{table}

\section{Synthetic Data Generation Prompt}
\label{sec:appendix_synth}

For generating synthetic test data (Section~4), we prompt proprietary models in a 5-shot fashion to generate text that resembles a given time period. The prompt instructs the model to produce a text snippet stylistically consistent with the target temporal class.

\begin{figure}[H]
\small
\begin{verbatim}
[System] You are an expert in historical linguistics. Generate a text passage that
is stylistically consistent with the specified time period. The passage should
reflect the vocabulary, syntax, and writing conventions typical of that era.

[User] Generate a Portuguese text passage (approximately 100 tokens) in the style
of the 1600s.

Here are some reference examples from that period:

[5 labeled examples from the 1600s class]

Generate a new text similar to these examples.
\end{verbatim}
\caption{Prompt template for synthetic test data generation. Reference examples are drawn from the training split of each dataset. This template is adapted for each corpus by adjusting the language, temporal class, and reference examples accordingly.}
\label{fig:synth_prompt}
\end{figure}

\end{document}